\documentclass{article}
\usepackage{spconf,amsmath,epsfig}
\usepackage[dvipsnames]{xcolor}
\usepackage[colorlinks=true,linkcolor=NavyBlue,citecolor=BrickRed]{hyperref}
\usepackage[capitalise]{cleveref}
\usepackage{array}
\usepackage{url}
\usepackage{verbatim}
\usepackage{graphicx}
\usepackage{multirow}
\usepackage{enumitem}

\title{Task Specific Pretraining with Noisy Labels\\ for Remote Sensing Image Segmentation}
\name{
  {Chenying Liu$^{1,2,3}$,\thanks{The work is to appear as a conference paper at IEEE IGARSS 2024. The work of C.\ Liu, C.\ M.\ Albrecht, and Y.\ Wang is funded by the Helmholtz Association through the Framework of \textit{HelmholtzAI}, grant ID: \texttt{ZT-I-PF-5-01} -- \textit{Local Unit Munich Unit @Aeronautics, Space and Transport (MASTr)}. The compute related to this work was supported by the Helmholtz Association's Initiative and Networking Fund on the HAICORE@FZJ partition. The work of X. X. Zhu is supported by the German Federal Ministry of Education and Research (BMBF) in the framework of the international future AI lab ``AI4EO -- Artificial Intelligence for Earth Observation: Reasoning, Uncertainties, Ethics and Beyond" (grant number: 01DD20001).\\
  }}
  {Conrad M Albrecht$^{2}$,}
  {Yi Wang$^{1}$,}
  {Xiao Xiang Zhu$^{1,3}$}}

\address{%
$^1$Data Science in Earth Observation, Technical University of Munich, Germany\\
$^2$Remote Sensing Technology Institute, German Aerospace Center, Germany\\
$^3$Munich Center for Machine Learning (MCML), Germany
}

\begin{document}
%
\maketitle
\begin{abstract}
Compared to supervised deep learning, self-supervision provides remote sensing a tool to reduce the amount of exact, human-crafted geospatial annotations. While image-level information for unsupervised pretraining efficiently works for various classification downstream tasks, the performance on pixel-level semantic segmentation lags behind in terms of model accuracy. On the contrary, many easily available label sources (e.g., automatic labeling tools and land cover land use products) exist, which can provide a large amount of noisy labels for segmentation model training. In this work, we propose to exploit noisy semantic segmentation maps for model pretraining. Our experiments provide insights on robustness per network layer. The transfer learning settings test the cases when the pretrained encoders are fine-tuned for different label classes and decoders. The results from two datasets indicate the effectiveness of task-specific supervised pretraining with noisy labels. Our findings pave new avenues to improved model accuracy and novel pretraining strategies for efficient remote sensing image segmentation.
\end{abstract}
\section{Introduction}
\label{sec:intr}

Deep learning turned into a powerful tool for data mining on vast amounts of remote sensing (RS) imagery \cite{zhu_deep_2017}. However, efficient training of deep learning models requires a large amount of accurate annotations, which is hard to obtain due to human labor intensive labeling process. Recently, self-supervised learning (SSL) has demonstrated great success in alleviating this problem by distillation of representative features from unlabeled data \cite{wang_self-supervised_2022}. Existing SSL methods such as contrastive learning \cite{caron_emerging_2021,he_momentum_2020} primarily rely on image-level information. Those turn out suboptimal for semantic segmentation downstream tasks relative to classification tasks \cite{wang_ssl4eo-s12_2023}. This discrepancy requests alternative strategies to enhance the efficacy of pretrained models for segmentation tasks. 

Recent studies indicate that deep learning models may be robust against label noise \cite{zhang_understanding_2021, liu_peaks_2022}. Models trained on billions of Instagram image--hashtag pairs without manual dataset curation exhibit excellent transfer learning performance for image classification tasks \cite{mahajan_exploring_2018}. Similar results were obtained when pretraining video models on large volumes of noisy social-media video data \cite{ghadiyaram_large-scale_2019}. In remote sensing, systematic studies have been carried out to employ crowd-sourced maps like OpenStreetMap (OSM) providing large-scale, publicly available labels for pretraining building and road extraction models \cite{kaiser_learning_2017, maggiori_convolutional_2017,zhang2020map}. Results indicate that OSM labels, though noisy, can significantly reduce human supervision required to successfully train segmentation models in these tasks.
Building upon the success of the existing works, we aim to further explore the potential of noisy labels in model pretraining for RS image segmentation tasks. We target to address the following questions:
\begin{enumerate}[leftmargin=5ex,topsep=2pt,itemsep=0.5ex,partopsep=0ex,parsep=0ex]
    \item Can supervised pretraining with noisy labels enhance the performance of encoders in general segmentation tasks compared to SSL methods? If so, what is the mechanism behind it?
    \item To what extent does the inconsistency of category definitions between pretraining and fine-tuning tasks impact the overall efficacy of the pretrained encoders?
    \item Are the encoders pretrained within a given framework useful when transferred to a different framework utilizing separate decoders for downstream tasks? 
\end{enumerate}
To answer these questions, we pretrain ResNet encoders in a supervised fashion on noisy labels to compare them with their SSL counterparts pretrained by DINO \cite{caron_emerging_2021} and MoCo \cite{he_momentum_2020}. We assemble two datasets to evaluate model effectiveness:
\begin{itemize}[leftmargin=5ex,topsep=2pt,itemsep=0.5ex,partopsep=0ex,parsep=0ex]
    \item the New York City (NYC) dataset representing a small-scale, in-domain scenario, and
    \item the SSL4EO-S2DW dataset potentially used for RS foundation model construction.
\end{itemize}
In the following, we first present the details of the two datasets and our experimental setups in \cref{sec:meth}, followed by results and corresponding discussions in \cref{sec:exp}. We summarize our findings for future lines in \cref{sec:conc}.

\section{Pretraining with Noisy Labels} \label{sec:meth}

\subsection{Datasets} \label{sec:meth:data}

\subsubsection{NYC dataset} \label{sec:meth:data:nyc}

\begin{figure}[t]
    \centering
    \footnotesize
    \begin{tabular}{p{2.cm}<{\centering}p{2.cm}<{\centering}p{2.cm}<{\centering}p{0.53cm}<{\centering}}
        \includegraphics[width=1.0\linewidth]{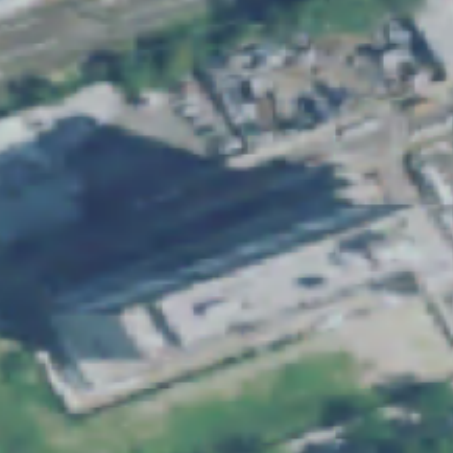} &  
        \includegraphics[width=1.0\linewidth]{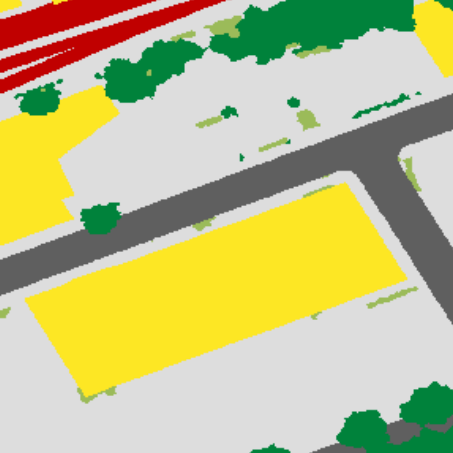} &
        \includegraphics[width=1.0\linewidth]{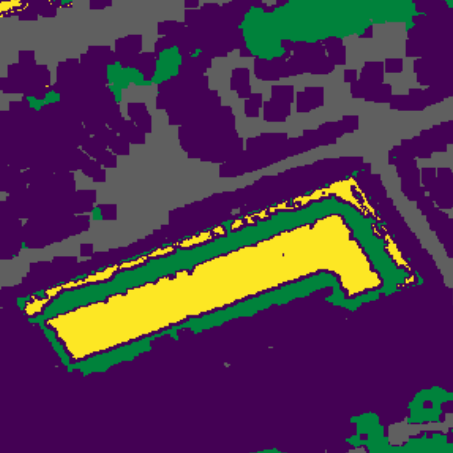} &
        \includegraphics[width=1.9\linewidth]{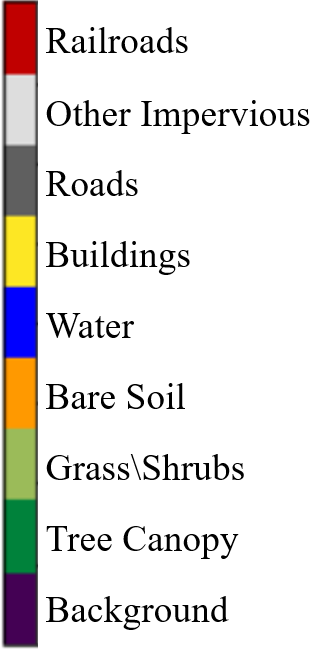} \\
        (a) Optical image & (b) GT & (c) Noisy labels
    \end{tabular}
    \caption{Visualization of a data triple within the NYC dataset.}
    \label{fig:data:nyc}
\end{figure}

The New York City (NYC) dataset was collected over New York City in 2017. We picked the 1m spatial resolution orthophotos as inputs. The four spectral bands are: near-infrared (NIR), red (R), green (G), and blue (B). We paired the pixel-level ground truth (GT) masks with 8 categories as depicted to the right of \cref{fig:data:nyc}. The noisy labels were generated from LiDAR data using the AutoGeoLabel approach as proposed in \cite{albrecht_autogeolabel_2021}, yet containing three classes, only: trees, buildings, and roads. Unclassified pixels are annotated as background. All data were curated into small patches of 288$\times$288 pixels, cf.\ \cref{fig:data:nyc}. A total of 26,500 data triples (orthophoto, GT, noisy labels) have been curated, with 4,500 hold out for testing. \Cref{tab:data:nycquality} quantifies the quality of the noisy labels.
We provide this dataset as test for the effectiveness of noisy label pretraining in a small-scale in-domain scenario. 22,000$-X$ data pairs serve pretraining either with or without noisy labels. The fine-tuning is implemented with 100 randomly selected image-GT pairs from the 22,000 pretraining patches. 

\begin{table}[t]
    \centering
    \footnotesize
    \caption{Quality assessment of the NYC noisy labels.}
    \label{tab:data:nycquality}
    \begin{tabular}{c|cccc|c}
         \hline\hline
         \textbf{CLASS} & \textbf{background} & \textbf{trees} & \textbf{buildings} & \textbf{roads} & \textbf{MEAN} \\
         \hline\hline
         OA & \multicolumn{5}{c}{67.83}\\
         \hline
         precision & 62.77 & 78.80 & 79.76 & 60.04 & 70.34 \\
         \hline
         recall    & 78.72 & 62.44 & 60.31 & 56.89 & 64.59 \\
         \hline
         IoU       & 53.67 & 53.46 & 52.30 & 41.26 & 50.17 \\
         \hline\hline
    \end{tabular}
\end{table}

\subsubsection{SSL4EO-S2DW dataset} \label{sec:meth:data:ssl4eo}

\begin{figure} [t]
    \centering
    \footnotesize
    \begin{tabular}{p{1.7cm}<{\centering}p{2cm}<{\centering}|p{1.7cm}<{\centering}p{1.7cm}}
        \includegraphics[width=1.15\linewidth]{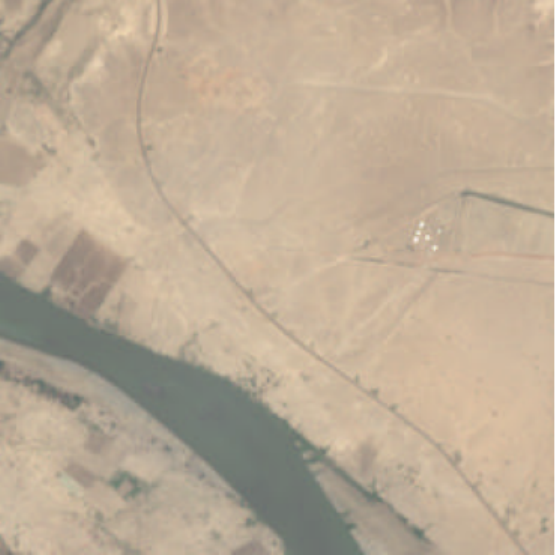} &  
        \includegraphics[width=0.97\linewidth]{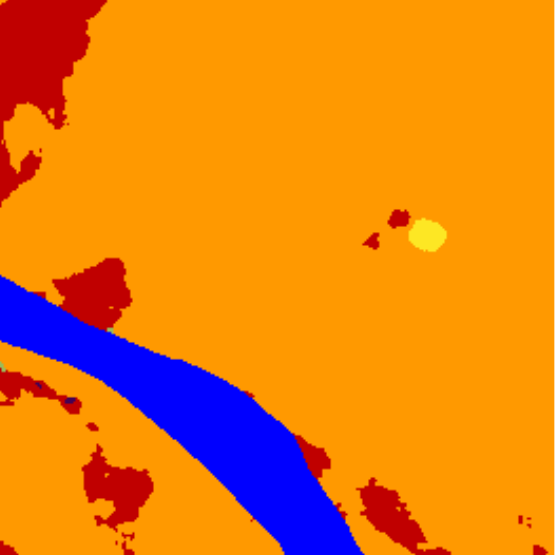} & 
        \includegraphics[width=1.15\linewidth]{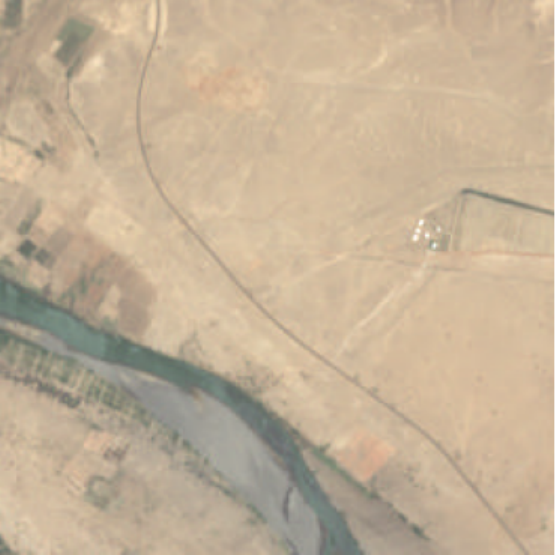} &
        \includegraphics[width=1.15\linewidth]{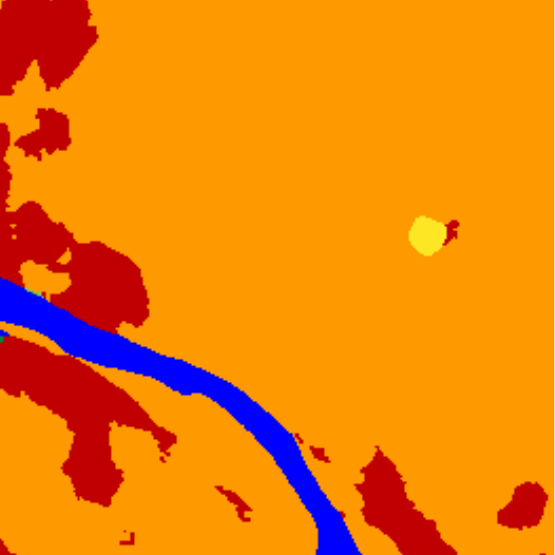} \\
        Optical image & Noisy labels & Optical image &  \hspace{0.3cm}Noisy labels
    \end{tabular}
    \caption{Visualization of orthophotos and corresponding noisy label masks for two seasons (left$\vert$right) at a random location of the SSL4EO-S2DW dataset. Blue, red, yellow, and orange represent water, crops, built area, and bare land, respectively.}
    \label{fig:data:ssl4eo}
\end{figure}

The dataset termed SSL4EO-S2DW we extend from the SSL4EO-S12 dataset \cite{wang_ssl4eo-s12_2023}, a large-scale self-supervision dataset for Earth observation. SSL4EO-S12 samples data globally from 251,079 locations. Each location corresponds to 4 Sentinel-1 and -2 image pairs of 264$\times$264 pixels from each season. Here, we only include Sentinel-2 data for SSL pretraining. We  pair them with the 9-class labels from the Google Dynamic World (DW) project \cite{brown_dynamic_2022}, cf.\ \cref{fig:data:ssl4eo}. The 9 classes include: water, trees, grass, flooded vegetation, crops, shrub and scrub, built area, bare land, and ice \& snow. We curated 103,793 locations with noisy label masks matching all seasons.
SSL4EO-S2DW resembles use cases where noisy labels are still a bit harder to obtain than abundant RS imagery. To evaluate pretrained encoders, we utilize the same downstream segmentation tasks as in \cite{wang_ssl4eo-s12_2023}, namely: DFC2020 \cite{yokoya_2020_2019} for land cover classification and OSCD for urban area change detection \cite{daudt_urban_2018}. We employ the training and test sets of the OSCD dataset. For the DFC2020 dataset, we utilize the 986 validation patches for fine-tuning, and 5128 test images for testing.

\begin{table*}[tp]
\centering
\footnotesize
\caption{Fine-tuning results (IoU, \%) obtained on the NYC dataset with different frameworks. }
\label{tab:ft:nyc}
\begin{tabular}{c|c|cccccccc|c}
\hline\hline
\textbf{Framework}                    & \textbf{Pretraining}  & \textbf{trees} & \textbf{grass/schrubs} & \textbf{bareland} & \textbf{water} & \textbf{buildings} & \textbf{roads} & \textbf{other impervious} & \textbf{railroads} & \textbf{mIoU}  \\ 
\hline\hline
                             & random       & 46.75 & 22.78         & \textbf{91.09}    & \textbf{96.55} & 44.69     & 32.73 & 30.67            & 90.86     & 54.28 \\ \cline{2-11} 
U-Net                        & DINO         & 49.34 & 22.90         & 79.38    & 90.04 & 46.15     & 42.31 & 31.70            & 91.14     & 57.09 \\ \cline{2-11} 
(fixed encoder)              & MoCo         & 47.56 & 22.65         & 78.76    & 72.13 & 47.21     & 42.27 & 31.50            & 91.10     & 56.52 \\ \cline{2-11} 
                             & noisy labels & \textbf{58.74} & \textbf{26.97}         & 91.05    & 81.74 & \textbf{59.37}     & \textbf{57.10} & \textbf{39.56}            & \textbf{91.20}     & \textbf{63.08} \\ 
\hline\hline
                             & random       & 48.39 & 19.03         & 79.11    & 86.24 & 48.53     & 43.20 & 29.02            & \textbf{90.26}     & 55.28 \\ \cline{2-11} 
DeepLabv3++                  & DINO         & 49.61 & 19.83         & 72.68    & 86.44 & 51.73     & 44.11 & 29.20            & 75.02     & 53.46 \\ \cline{2-11} 
(fine-tuned encoder)            & MoCo         & 49.13 & 20.65         & 69.05    & 87.41 & 51.98     & 47.82 & 30.24            & 82.68     & 54.76 \\ \cline{2-11} 
                             & noisy labels & \textbf{54.78} & \textbf{23.86}         & \textbf{84.41}    & \textbf{92.08} & \textbf{59.22}     & \textbf{58.04} & \textbf{37.99}            & 81.10     & \textbf{61.31} \\ 
\hline\hline
\end{tabular}
\end{table*}

\subsection{Implementation Details} \label{sec:meth:imp}

For pretraining, we use image data and noisy label pairs to train U-Nets with ResNet encoder backbones in a standard supervised setup. Given the dataset sizes, we chose ResNet18 (NYC) and ResNet50 (SSL4EO-S2DW). For transfer learning, we test the pretrained encoders within different frameworks: U-Net \cite{ronneberger_u-net_2015}, DeepLabv3++ \cite{chen_encoder-decoder_2018}, and PSPNet \cite{zhao_pyramid_2017}. Our pretrained encoders are compared with random initialization and those obtained by DINO and MoCo from \cite{wang_ssl4eo-s12_2023}.

We applied an Adam optimizer on a loss combining CrossEntropy and Dice. Random flipping served as our data augmentation strategy. The pretraining learning rate we set to \texttt{1e-3}. We use a smaller learning rate of \texttt{5e-4} adjusted by a $\cos$-scheduler for fine-tuning. For SSL4EO-S2DW pretraining, we randomly cropped patches into 256$\times$256, and chose the data from an arbitrary season at each geospatial location and training iteration to act as an additional augmentation strategy. We pretrain the models with a batch size of 256 per GPU. Pre-training for 100 epochs takes about 5 hours on an NVIDIA A100 GPU with the NYC dataset. For SSL4EO-S2DW it takes 4 GPUs to train for 100 epochs in 6 hours.

\begin{table}[t]
\centering
\footnotesize
\caption{Fine-tuning results (\%) obtained on the DFC2020 dataset using PSPNet as frameworks, where OA presents overall accuracy, and AA is average accuracy.}
\label{tab:ft:dfc}
\begin{tabular}{c|ccc|ccc}
\hline\hline
\multirow{2}{*}{Pretraining}       & \multicolumn{3}{c|}{fixed encoder} & \multicolumn{3}{c}{fine-tuned encoder} \\
\cline{2-7}
             & OA    & mIoU    & AA  & OA    & mIoU    & AA    \\
\hline\hline
random       & 56.42 & 31.50 & 45.12  & 58.68 & 33.56   & 46.03  \\
\hline
DINO         & 64.82 & 37.81 & 48.83  & 63.64 & 36.95   & 49.92  \\
\hline
MoCo         & 63.25 & 37.67 & 51.00  & 61.19 & 34.86   & 47.29  \\
\hline
noisy labels & \textbf{66.66} & \textbf{40.88} & \textbf{53.24}  & \textbf{67.11} & \textbf{41.06}   & \textbf{53.14}  \\
\hline\hline
\end{tabular}
\end{table}

\section{Experimental Results}
\label{sec:exp}


\subsection{Transfer Learning} \label{sec:exp:results}

\subsubsection{NYC dataset} \label{sec:exp:results:nyc}

We transfer the (3+1)-class noisy label pretrained encoder to an 8-class land cover land use segmentation downstream task. While we freeze the encoder when the downstream task utilizes the same framework as in pretraining (U-Net), we let adjust the encoder weights along with the decoder when adopting a different framework (DeepLabv3++). As shown in \cref{tab:ft:nyc}, the noisy label pretrained encoder outperforms the other models on almost all classes although the pretrained model has not been pretrained on some of the classes: Including semantic information for pretraining is beneficial for models to learn generic features that are discriminative for semantic segmentation downstream tasks. Notably, the pretrained encoder works for different training frameworks, too. In this case, pretrained encoders seem compatible when transferred to, e.g., DeepLabv3++. In contrast, the two SSL methods fail to show an edge over random initialization on the NYC dataset. We partly attribute this result to a lack of large amounts of unlabeled data in a small-scale setup. 

\subsubsection{SSL4EO-S2DW dataset} \label{sec:exp:results:ssl4eo}

We picked PSPNet and U-Net as frameworks for the DFC2020 and OSCD datasets. We test two fine-tuning settings with fixed and fine-tuned encoders on both datasets. \Cref{tab:ft:dfc} and \cref{tab:ft:oscd} present results, respectively. 
Noisy label--pretrained encoders yield better results when compared to SSL--pretraining or when referenced to random initialization. Performance margins increase when the encoders are fixed in the fine-tuning stage, which indicates that the encoder pretrained with noisy labels is able to generate features adapted to segmentation tasks. Our experiments on two distinct downstream tasks further illustrate the generalizability of encoders pretrained by noisy labels.  

\begin{table}[t]
\centering
\footnotesize
\caption{Fine-tuning results (\%) obtained on the OSCD dataset using U-Net as frameworks.}
\label{tab:ft:oscd}
\begin{tabular}{c|ccc|ccc}
\hline\hline
\multirow{2}{*}{Pretraining}       & \multicolumn{3}{c|}{fixed encoder} & \multicolumn{3}{c}{fine-tuned encoder} \\
\cline{2-7}
             & OA    & IoU    & Precision  & OA    & IoU    & Precision    \\
\hline\hline
random       & 95.47 & 17.08 & \textbf{78.06}  & 95.16 & 21.80   & 66.27  \\
\hline
DINO         & 95.59 & 21.74 & 73.83  & 95.53 & 31.05   & 66.45  \\
\hline
MoCo         & 95.66 & 23.81 & 73.34  & 95.70 & 32.56   & 66.39  \\
\hline
noisy labels & \textbf{95.79} & \textbf{26.80} & 73.90  & \textbf{95.98} & \textbf{33.37}   & \textbf{71.34}  \\
\hline\hline
\end{tabular}
\end{table}

\begin{figure*}[t]
    \centering
    \includegraphics[width=1.\linewidth]{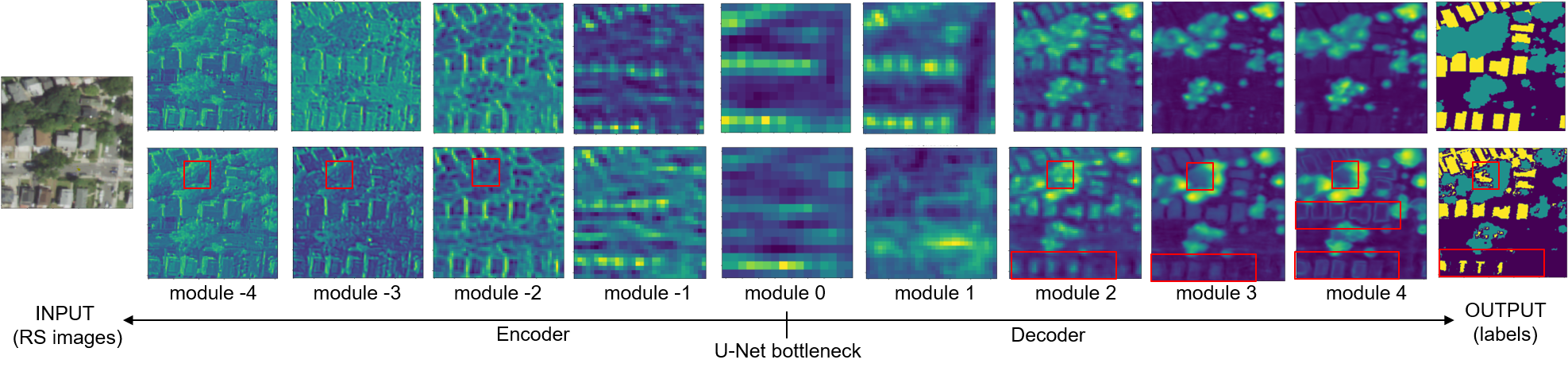}
    \caption{Visualization of the dominant principle component from the output of each convolutional layer of the U-Net model trained with exact labels (top row) and noisy labels (bottom row).}
    \label{fig:nyc:feat}
\end{figure*}

\subsection{Impact of Label Noise on Model Training} \label{sec:exp:data}

To understand the mechanism behind the success of noisy label pretraining, we utilize a 3-class version of the NYC dataset discriminating trees, buildings, and background as illustrated in \cref{fig:nyc:feat}. We train two U-Net models from scratch employing all available training patches; one with noisy labels and one with exact labels (GT masks), respectively. After training, we analyze the output features of each convolutional layer of the U-Net given an input patch. For one input patch, we visualize the first principle component of each such output feature data cube in \cref{fig:nyc:feat}. We observe:
\begin{itemize}[leftmargin=5ex,topsep=2pt,itemsep=0.5ex,partopsep=0ex,parsep=0ex]
    \item the encoder features visually share spatial characteristics, i.e., the encoder seems little impacted by label noise
    \item the closer the convolutional layer gets to the U-Net's output, the more the features become contaminated by label noise
\end{itemize}
Since the encoder learns to extract basic spatial features from local semantic information of the input data, it is affected little by label noise. In terms of backpropagation, decoders are closer to the (noisy) label mask to optimize the U-Net's output on. Thus, while the decoder adapts to output noisy labels, the encoder is less biased by label noise. 

\begin{figure}[t]
    \centering
    \footnotesize
    \begin{tabular}{m{3.7cm}<{\centering}m{3.7cm}<{\centering}}
        \includegraphics[width=1.08\linewidth]{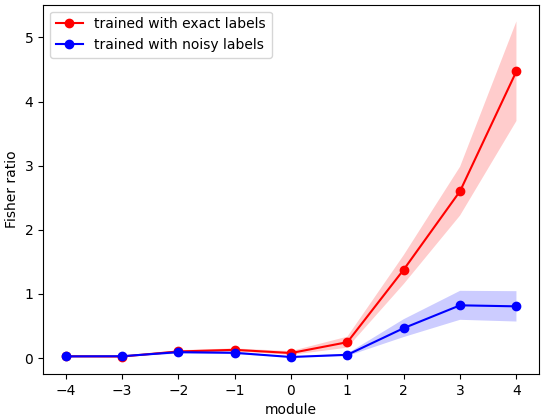} & 
        \includegraphics[width=1.13\linewidth]{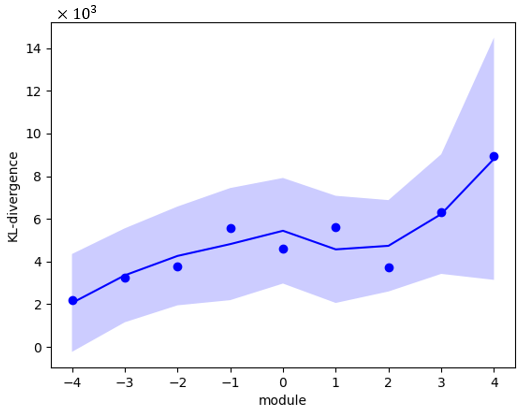} \\
        (a) Fisher ratios & (b) KL divergences
    \end{tabular}
    \caption{Quantitative assessment of data statistics for convolutional modules after U-Net training: (a) Fisher ratios of module output features after U-Net training. Shaded areas indicate standard deviations w.r.t.\ data samples investigated. (b) KL divergence of module weight statistics comparing the model trained with exact labels to the one trained on noisy labels. The solid line connects the smoothed results by a Savitzky–Golay filter, and the shaded areas indicates the standard deviation independently training 5 models from scratch.}
    \label{fig:nyc:fldkl}
\end{figure}


To quantify our observations, we calculate the Fisher ratios of output feature cubes as presented in \cref{fig:nyc:fldkl} (a). The Fisher ratio is a widely used index to assess feature discrimination in pattern recognition \cite{liu_feature-driven_2018}: Larger values indicate better discrimination of features. Two insights we read from \cref{fig:nyc:fldkl}:
\begin{enumerate}[leftmargin=5ex,topsep=2pt,itemsep=0.5ex,partopsep=0ex,parsep=0ex]
    \item decoder features, whether trained on noisy or exact labels, are more discriminative compared to encoder features
    \item the discriminative character of decoder features is degraded when the model is trained on noisy labels, i.e., the decoder is significantly affected by label noise
\end{enumerate}
We did also investigate the model training with exact and noisy labels by computing the Kullback–Leibler (KL) divergence \cite{mackay_information_2003} of weight statistics within each module. As demonstrated in \cref{fig:nyc:fldkl} (b), the convolutional layers in the encoder are governed by similar weight statistics, while those in the decoder follow diverging weight statistics towards the semantic segmentation outputs. Those observations highlight that encoder features are less biased by label noise, yet, they benefit from the semantics provided by pixel-level noisy label masks. 


\section{Conclusions \& Perspectives}
\label{sec:conc}

In this work, we explored the effectiveness of pixel-wise noisy labels for pre-training deep artificial neural networks on semantic segmentation downstream tasks. Experiments utilizing two remote sensing datasets demonstrate the potential of the proposed approach in order to enhance the performance of encoder networks for various segmentation downstream tasks---even when the class definitions or the training frameworks of fine-tuning divert from the pretraining setup. Furthermore, we tapped on explaining why encoders seem robust against label noise for U-Net models. 
Our findings motivate future work to test the pretrained encoders on a more diverse set of downstream tasks. Additional experiments beyond U-Net training will help to solidify our initial observations.

\bibliographystyle{IEEEbib}
\bibliography{refs}

\end{document}